\pdfoutput=1
\documentclass{article}

\PassOptionsToPackage{numbers, compress}{natbib}
\usepackage[preprint]{neurips_2023}

\usepackage[utf8]{inputenc} % allow utf-8 input
\usepackage[T1]{fontenc}    % use 8-bit T1 fonts
\usepackage{hyperref}       % hyperlinks
\usepackage{url}            % simple URL typesetting
\usepackage{booktabs}       % professional-quality tables
\usepackage{amsfonts}       % blackboard math symbols
\usepackage{nicefrac}       % compact symbols for 1/2, etc.
\usepackage{microtype}      % microtypography
\usepackage{xcolor}         % colors

\usepackage{graphicx}
\usepackage{multicol, multirow}
\usepackage{caption}

\usepackage{amsmath,amsfonts,bm}

\def\eqref#1{equation~\ref{#1}}
\def\1{\bm{1}}

\def\rvt{{\mathbf{t}}}

\def\rvx{{\mathbf{x}}}
\def\rvy{{\mathbf{y}}}

\def\rmA{{\mathbf{A}}}

\def\rmQ{{\mathbf{Q}}}
\def\rmR{{\mathbf{R}}}
\def\rmS{{\mathbf{S}}}
\def\rmT{{\mathbf{T}}}

\def\rmV{{\mathbf{V}}}

\def\rmX{{\mathbf{X}}}

\DeclareMathAlphabet{\mathsfit}{\encodingdefault}{\sfdefault}{m}{sl}
\SetMathAlphabet{\mathsfit}{bold}{\encodingdefault}{\sfdefault}{bx}{n}

\def\gI{{\mathcal{I}}}

\def\gN{{\mathcal{N}}}

\def\gU{{\mathcal{U}}}

\def\sN{{\mathbb{N}}}

\def\sR{{\mathbb{R}}}

\DeclareMathOperator*{\argmin}{arg\,min}

\newcommand{\SOthree}{\mathrm{SO}(3)}
\newcommand{\SEthree}{\mathrm{SE}(3)}

\newcommand{\Sym}{\mathfrak{S}}
\newcommand{\hrvy}{\hat{\rvy}}

\newcommand{\wrt}{\textit{w.r.t.}~}
\newcommand{\st}{\textit{s.t.}~}

\renewcommand{\paragraph}[1]{\vspace{0em}\noindent\textbf{#1}.}
\usepackage{color}
\usepackage{xcolor}
\definecolor{turquoise}{cmyk}{0.65,0,0.1,0.3}
\definecolor{purple}{rgb}{0.65,0,0.65}
\definecolor{dark_green}{rgb}{0, 0.5, 0}
\definecolor{orange}{rgb}{0.8, 0.6, 0.2}
\definecolor{red}{rgb}{0.8, 0.2, 0.2}
\definecolor{darkred}{rgb}{0.6, 0.1, 0.05}
\definecolor{blueish}{rgb}{0.0, 0.3, .6}
\definecolor{light_gray}{rgb}{0.7, 0.7, .7}
\definecolor{pink}{rgb}{1, 0, 1}
\definecolor{greyblue}{rgb}{0.25, 0.25, 1}
\definecolor{tab_blue}{HTML}{1f77b4}
\definecolor{tab_orange}{HTML}{ff7f0e}
\definecolor{LightRed}{rgb}{0.99,0.89,0.89}

\definecolor{mesh_misty_rose}{HTML}{e6aaa3}
\definecolor{mesh_yellow}{HTML}{ffba00}

\newcommand{\SupplementaryMaterial}{{\color{darkred} supplementary material}}

\title{\emph{Banana}: \emph{Bana}ch Fixed-Point \emph{N}etwork for Pointcloud Segmentation with Inter-Part Equiv\emph{a}riance}

\author{%
  Congyue Deng\textsuperscript{1}$^*$ \quad Jiahui Lei\textsuperscript{2}$^*$ \quad Bokui Shen\textsuperscript{1} \quad  Kostas Daniilidis\textsuperscript{2} \quad Leonidas Guibas\textsuperscript{1} \\
  $^1$ Stanford University \qquad
  $^2$ University of Pennsylvania \\
  {\tt\small \{congyue, willshen, guibas\}@cs.stanford.edu, \{leijh, kostas\}@cis.upenn.edu}
}

\begin{document}

\maketitle
\def\thefootnote{*}\footnotetext{Equal contribution.}

\begin{abstract}
    Equivariance has gained strong interest as a desirable network property that inherently ensures robust generalization.
    However, when dealing with complex systems such as articulated objects or multi-object scenes, effectively capturing inter-part transformations poses a challenge, as it becomes entangled with the overall structure and local transformations. The interdependence of part assignment and per-part group action necessitates a novel equivariance formulation that allows for their co-evolution.
    In this paper, we present \textit{Banana}, a Banach fixed-point network for pointcloud segmentation with inter-part equivariance \textbf{by construction}.
    Our key insight is to iteratively solve a fixed-point problem, where point-part assignment labels and per-part $\SEthree$-equivariance co-evolve simultaneously.
    We provide theoretical derivations of both per-step equivariance and global convergence, which induces an equivariant final convergent state.
    Our formulation naturally provides a strict definition of inter-part equivariance that generalizes to unseen inter-part configurations.
    Through experiments conducted on both articulated objects and multi-object scans, we demonstrate the efficacy of our approach in achieving strong generalization under inter-part transformations, even when confronted with substantial changes in pointcloud geometry and topology.
\end{abstract}
\section{Introduction}

From articulated objects to multi-object scenes, multi-body systems, i.e. shapes composed of multiple parts where each part can be moved separately, are prevalent in various daily-life scenarios. 
However, modeling such systems presents considerable challenges compared to rigid objects, primarily due to the infinite shape variations resulting from inter-part pose transformations with exponentially growing complexities. Successfully modeling these shapes demands generalization across potential inter-part configurations.
While standard data augmentation techniques can potentially alleviate such problems, exhaustive augmentation can be highly expensive especially given the complexity of such systems compared to rigid objects.
Meanwhile, recent rigid-shape analysis techniques have made significant progress in building generalization through the important concept of equivariance \cite{kondor2018generalization, cohen2019general, weiler2021coordinate, aronsson2022homogeneous, xu2022unified, thomas2018tensor, fuchs2020se, chen2021equivariant, deng2021vector, assaad2022vn, katzir2022shape, li2022directed, poulenard2021functional}. Equivariance dictates that when a transformation is applied to the input data, the network's output should undergo a corresponding transformation. For example, if a point cloud is rotated by 90 degrees, the segmentation masks should also rotate accordingly. However, to the best of our knowledge, no existing work has managed to extend the formulation of equivariance to inter-part configurations. 

In this paper, we address the challenge of achieving inter-part equivariance in the context of part segmentation, a fundamental task in point cloud analysis, which is also the key to generalizing equivaraince from single object to multi-body system. Extending equivariance to multi-body systems presents notable challenges in both formulation and realization. When dealing with a shape that allows for inter-part motions, the model must exhibit equivariance to both global and local state changes, which can only be defined by combining part assignment and per-part transformations. For example, to model an oven's inter-part states, we first need to differentiate between the parts corresponding to the door and the body and then define the movement of each individual part.
An intriguing "chicken or the egg" problem arises when attempting to build inter-part equivariance without access to a provided segmentation, where segmentation is necessary to define equivariance, but segmentation itself is the desired output that we aim to produce.

Our key insight to tackle this seeming dilemma is to model inter-part equivariance as a sequential fixed-point problem by co-evolving part segmentation labels and per-part $\SEthree$-equivariance. We provide theoretical derivations for both the per-step behavior and global convergence behavior, which are crucial aspects of a fixed-point problem. We show that our formulation establishes per-step equivariance through network construction, which then induces an overall inter-part equivariance upon convergence. Thus, by having equivariant per-step progression and global convergence, our formulation naturally gives rise to a strict definition of inter-part equivariance through iterative inference that generalizes to unseen part configurations.
We further bring our formulation to concrete model designs by proposing a novel part-aware equivariant network with a weighted message-passing paradigm. With localized network operators and per-part $\SEthree$-equivaraint features, the network is able to guarantee per-step inter-part equivariance as well as facilitate stable convergence during the iterative inference.
We test our framework on articulated objects to generalize from static rest states to novel articulates and multi-object scenes to generalize from clean synthetic scenes to cluttered real scans. Our model shows strong generalization in both scenarios even under significant changes in pointcloud geometry or topology.

To summarize, our key contributions are
\begin{itemize}
    \item To the best of our knowledge, we are the first to provide a strict definition of inter-part equivariance for pointcloud segmentation and introduce a learning framework with such equivariance \textbf{by construction}.
    \item We propose a fixed-point framework with one-step training and iterative inference and show that the per-step equivariance induces an overall equivariance upon convergence.
    \item We design a part-aware equivariant message-passing network with stable convergence. 
    \item Experiments show our strong generalization under inter-part configuration changes even when they cause subsequent changes in pointcloud geometry or topology.
\end{itemize}

\begin{figure*}[t]
    \centering
    \includegraphics[width=0.99\linewidth]{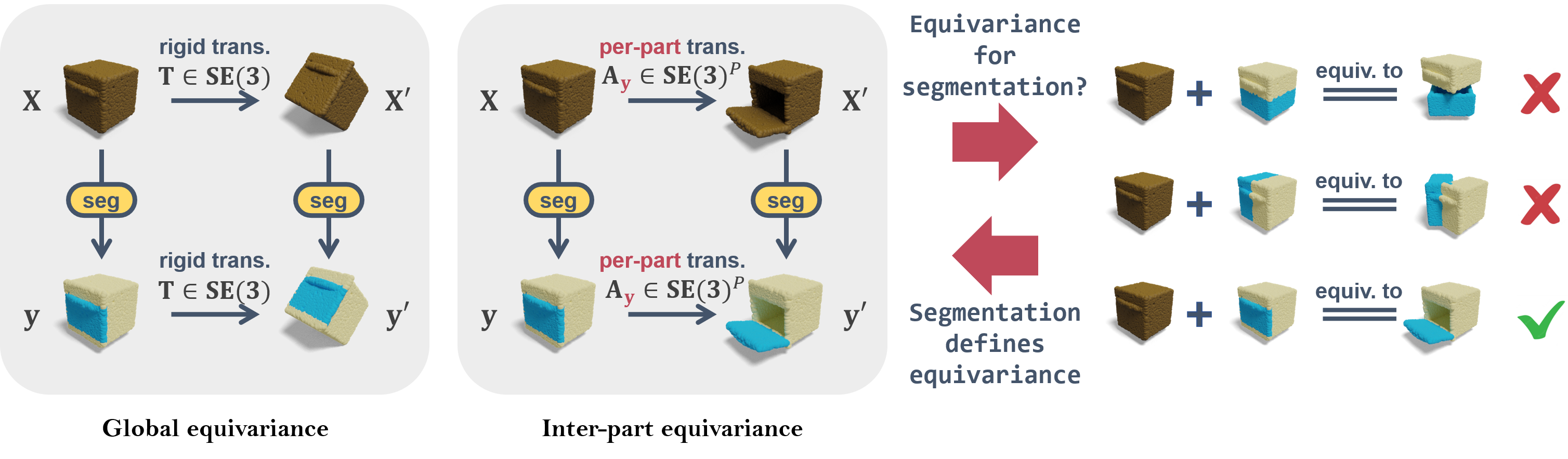}
    \caption{\textbf{"Chicken or the egg" problem with inter-part equivariance.}
    \textbf{Left:} Rigid $\SEthree$-equivariance. If a rigid transformation is applied to the input pointcloud, the output segmentation transforms accordingly.
    \textbf{Right:} Inter-part equivariance. Analogously, we want the network input and output to be coherent under \emph{per-part} transformations. But such a definition of equivariance requires part segmentations, which is also exactly the desired network output, resulting in a "chicken or the egg" problem.
    }
    \label{fig:teaser}
\end{figure*}
\section{Related Work}

\paragraph{Equivariant pointcloud networks}
Existing works in equivariant 3D learning mainly focus on rigid $\SEthree$ transformations.
As a well-studied problem, it has developed comprehensive theories \cite{kondor2018generalization, cohen2019general, weiler2021coordinate, aronsson2022homogeneous, xu2022unified} and abundant network designs \cite{thomas2018tensor, fuchs2020se, chen2021equivariant, deng2021vector, assaad2022vn, katzir2022shape, li2022directed, poulenard2021functional}, which benefit a variety of 3D vision and robotics tasks ranging from pose estimation \cite{li2021leveraging, lin2022coarse, pan2022so, sajnani2022condor, zhu2022correspondence}, shape reconstruction \cite{chen2021equivariant, chatzipantazis2022se}, to object interaction and manipulation \cite{fu2022robust, higuera2022neural, ryu2022equivariant, simeonov2022neural, weng2022neural, xue2022useek}.
A few recent works have extended $\SEthree$ equivariance to part level \cite{yu2022rotationally, lei2023efem, liu2023self}.
\cite{yu2022rotationally} employs a local equivariant feature extractor for object bounding box prediction, showing robustness under object-level and scene-level pose changes.
\cite{lei2023efem} learns an $\SEthree$-equivaraint object prior and applies it to object detection and segmentation in scenes with robustness to scene configuration changes.
\cite{liu2023self} learns part-level equivariance and pose canonicalization from a collection of articulated objects with a two-stage coarse-to-fine network.
However, all these works are purely heuristics-based, and none has provided a strict definition of inter-part equivariance.

\paragraph{Multi-body systems}
From small-scale articulated objects to large-scale multi-object scenes, there is a wide range of 3D observations that involve multiple movable parts or objects, exhibiting various configurations and temporal variations. This has sparked a rich variety of works dedicated to addressing the challenges in multi-body systems, studying their part correspondences \cite{puy2020flot,liu2019meteornet,behl2018pointflownet,liu2019flownet3d,li2021self}, segmentations \cite{yi2018deep,wang20224d,song2022ogc,huang2022dynamic,huang2021multibodysync,thomas2021self,baur2021slim,chen20224dcontrast}, reconstructions \cite{jiang2022ditto, kawana2021unsupervised,jain2021screwnet, mu2021sdf, Lei2022CaDeX}, or rearrangements \cite{wei2023lego,tang2023diffuscene}, to name just a few.
However, as we know the system is acted via products of group actions; a more structured network that exploits such motion prior by construction is desired. Only a few works have studied this problem.
\cite{yu2022rotationally} exploits global and local gauge equivariance in 3D object detection via bottom-up point grouping but without a systematical study of inter-object transformations.
\cite{lei2023efem} further exploits equivariance of object compositions in scene segmentation, but as an EM-based approach, it requires exhaustive and inefficient enumerations and no global convergence is guaranteed.
In contrast, our approach offers a theoretically sound framework that achieves multi-body equivariance by construction, which enables robust analysis for various tasks and provides a more structured and reliable solution compared to existing methods.

\paragraph{Iterative inference}
From recurrent neural networks \cite{goller1996learning, hochreiter1997long} to flow \cite{dinh2014nice, rezende2015variational, dinh2016density, kingma2018glow, yang2019pointflow} or diffusion-based generative models \cite{song2019generative, song2020score, ho2020denoising, song2020denoising, luo2021diffusion, zhou20213d}, iterative inference has played many diverse yet essential roles in the development of computer vision.
Though naturally adapted to sequential or temporal data, iterative inference can also be applied to static learning problems such as image analysis \cite{girshick2014rich, girshick2015fast, wang2016cnn, he2017mask} and pointcloud segmentation \cite{ye20183d, liu20173dcnn}.
Recent studies also show that iterative inference presents stronger generalizability than one-step forward predictions \cite{schwarzschild2021can, bansal2022end}, with explanations of their relations to the "working memory" of human minds \cite{baddeley1992working, baddeley2012working} or human visual systems \cite{liao2016bridging, kar2019evidence}.
In our work, we employ iterative inference on static pointclouds to co-evolve two intertwined attributes: part segmentation and inter-part equivaraince, in order to address a seemingly contradictory situation.

\section{Inter-Part Equivariance}
\label{sec:method1:art_equiv}

\begin{figure*}[t]
    \centering
    \includegraphics[width=0.99\linewidth]{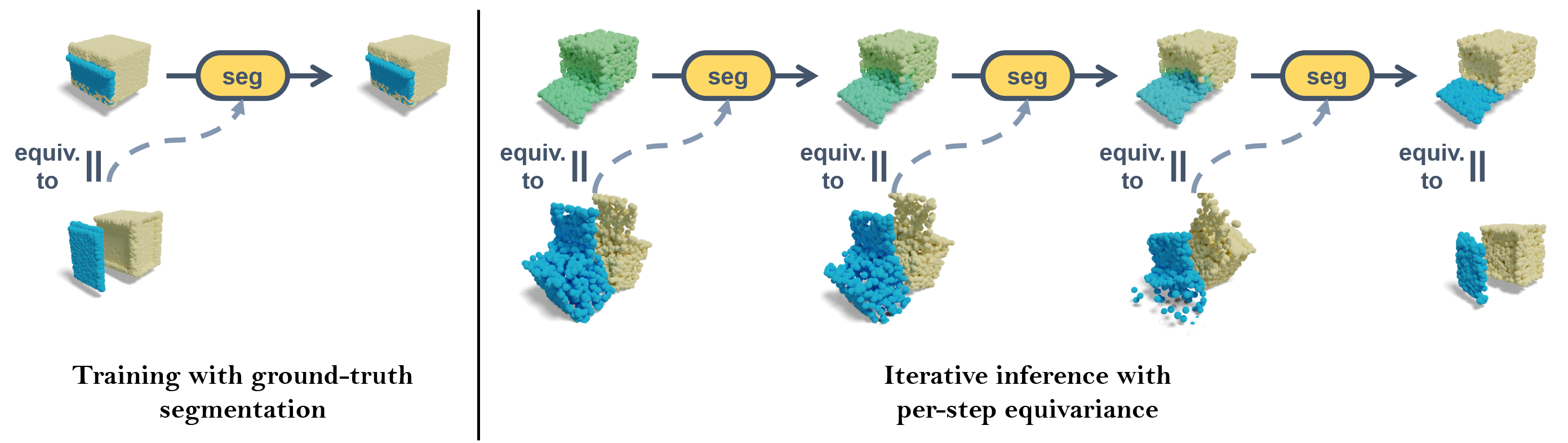}
    \caption{\textbf{Single-state training and novel-state iterative inference.}
    \textbf{Left:} Training with ground-truth segmentation as both network input and output.
    \textbf{Right:} Equivariant Banach iterations.
    At each step $k$, the network is equivariant to its current segmentation input $\rvy^{(k)}$ by construction. At the end of the iterations, it converges to its fixed point with an induced overall equivariance.
    }
    \label{fig:method_iters}
\end{figure*}

We start by defining inter-part equivariance on pointclouds with given part segmentations (Sec. \ref{sec:method1:art_group}).
Then, we will introduce how to extend this definition to unsegmented pointclouds using a fixed-point framework (Sec. \ref{sec:method1:banach_fp}) with one-step training and iterative inference.
The iterations have per-step equivariance by network construction, which is shown to induce an overall equivariance upon convergence (Sec \ref{sec:method1:fp_equiv}).
Finally, we will also explain how to eliminate part orders for instance segmentation (Sec \ref{sec:method1:part_permut}).

\subsection{Multi-Body Systems}
\label{sec:method1:art_group}

Let $\rmX\in\sR^{N\times3}$ be a pointcloud with $P$ parts. Segmentations on $\rmX$ are represented as point-part assignment matrices $\rvy \in [0, 1]^{N\times P}$ which sum up to 1 along the $P$ dimension.
As the action of $\SEthree^P$ on $\rmX$ and its resulting equivariance is subject to the part decompositions, we define the action of $\SEthree^P$ on (pointcloud, segmentation) pairs $(\rmX, \rvy)$.
When $\rvy$ is a binary mask, the $P$ parts of $\rmX$ are $P$ disjoint sub-pointclouds and each $\SEthree$ component of $\SEthree^P$ acts on one part separately.
More concretely, for any transformation $\rmA = (\rmT_1, \cdots, \rmT_P) \in \SEthree^P$, we define $\rmA(\rmX,\rvy) := (\rmX', \rvy)$ where the $n$-th point $\rvx_n$ is transformed to $\rvx'_n$ by
\begin{equation}
    \rvx'_n := \sum_{p=1}^P \rvy_{np} (\rvx_n\rmR_p + \rvt_p).
\end{equation}
where $\rmR_p$ and $\rvt_p$ are the rotation and translation components of $\rmT_p\in\SEthree$. For soft segmentation masks, we view each point as probabilistically assigned to one of the $P$ parts.

\paragraph{Equivariance}
Similar to $\SEthree$-equivariance on rigid shapes $f(\rmT\rmX) = \rmT f(\rmX), \forall\rmT\in\SEthree$, we define the inter-part equivariance on $(\rmX,\rvy)$ pairs by saying that a function $f$ is $\SEthree^P$-equivariant if $\forall\rmA \in \SEthree^P$,
\begin{equation}
    f(\rmA(\rmX,\rvy)) = \rmA f(\rmX,\rvy).
\end{equation}
Note that if the function outputs $f(\rmX,\rvy)$ are no longer (pointcloud, segmentation) pairs, the $\SEthree^P$-action on its outputs needs to be specified.
A special case is when $\SEthree^P$ acts trivially on $f(\rmX,\rvy)$ by $f(\rmA(\rmX,\rvy)) = f(\rmX,\rvy)$, which defines the common-sense ``invariance''.
For a pointcloud segmentation network with per-point part-label outputs $\rvy\in[0,1]^{N\times P}$, we desire it to be $\SEthree^P$-invariant under the above definitions, that is $f(\rmA(\rmX,\rvy)) \equiv \rvy, \,\forall \rmA\in\SEthree^P$.

\subsection{Banach Fixed-Point Iterations}
\label{sec:method1:banach_fp}

\paragraph{``Chicken or the egg''}
The "chicken or the egg" nature of segmentation with inter-part equivariance immediately emerges from the above definition: to enforce $\SEthree^P$-equivariance for a neural network $f(\cdot;\Theta)$, the part segmentation $\rvy$ is required as input, which is also exactly the desired output of the network (Fig. \ref{fig:teaser}). That this,
\begin{equation}
\label{eq:chicken_egg}
    f(\rmX, \rvy; \Theta) = \rvy.
\end{equation}
To resolve this dilemmatic problem, we approach Eq. \ref{eq:chicken_egg} as a fixed-point equation on function $f(\cdot, \rvy; \Theta)$ \wrt $\rvy$.
And instead of using single-step forward predictions, we solve it with Banach fixed-point iterations (Fig. \ref{fig:method_iters}).

\paragraph{Training}
At training time, we aim to optimize the network weights $\Theta$ such that the labeled point clouds $(\rmX_i,\rvy_i), i\in\gI$ from the dataset become fixed points of the function $f(\cdot,\Theta)$:
\begin{equation}
\label{eq:training}
    \Theta_* = \argmin_\Theta \frac{1}{|\gI|} \sum_{i\in\gI} \|f(\rmX_i, \rvy_i; \Theta) - \rvy_i\|.
\end{equation}
However, there exists a trivial solution which is the identity function $f(\cdot,\rvy;\Theta) \equiv \rvy$, meaning that any arbitrary segmentation can satisfy the objective for any pointclouds. Thus, to avoid such degenerated cases, we have to limit the network expressivity. More specifically, we limit the Lipschitz constant $L$ of the network \wrt $\rvy$ which is defined by
\begin{equation}
    \|f(\rmX,\rvy_1;\Theta) - f(\rmX,\rvy_2;\Theta)\| \leq L \|\rvy_1 - \rvy_2\|, \quad
    \forall\rmX, \, \forall \rvy_1,\rvy_2.
\end{equation}
If $f$ degenerates to the identical mapping, its Lipschitz $L = 1$, which violates the above constraint.
When $L < 1$, $f$ is a contractive function and has a unique fixed point based on the Banach fixed-point theorem. Our training objective is to align this fixed point with the ground-truth segmentation.

\paragraph{Iterative inference}
At inference time, given an input pointcloud $\rmX_0$, the learnable parameters $\Theta_*$ fixed in $f$ and we look for a segmentation $\rvy$ that satisfies
\begin{equation}
    f_{\rmX_0,\Theta_*}(\rvy) := f(\rmX_0, \rvy; \Theta_*) = \rvy.
\end{equation}
We solve this equation with Banach fixed-point iterations \cite{agarwal2018banach}
\begin{equation}
\label{eq:banach_iter}
    \rvy^* = \lim_{k\to\infty} \rvy^{(k)} = \lim_{k\to\infty} f^{(k)}_{\rmX_0,\Theta_*}(\rvy^{(0)})
\end{equation}
where $\rvy^{(0)}$ is a random initialization of the segmentation on $\rmX_0$.
For a contractive function $f$, the uniqueness of $\rvy^*$ induces a well-defined mapping from pointclouds to segmentations $f_\Theta^*: \rmX_0 \to \rvy^*$.

\subsection{Equivariance}
\label{sec:method1:fp_equiv}

For an input pairs $(\rmX,\rvy)$ with known segmentation $\rvy$, we employ an $\SEthree$-equivariant backbone \cite{deng2021vector} on each individual part and construct an $\SEthree^P$-equivariant network, which will be further explained in Section \ref{sec:method2:network}.
This guarantees the inter-part equivariance at training time and at each timestep during inference, but most importantly we desire the overall $\SEthree^P$-equivariance at the convergence point of the Banach iteration. Here we prove two properties:

\paragraph{Self-coherence of convergent state}
At each inference timestep $k$, we have $\SEthree^P$-equivariance subject to $\rvy^{(k)}$, that is, $\forall\rmA\in\SEthree^P$,
\begin{equation}
    f(\rmA(\rmX_0,\rvy^{(k)}); \Theta) = f(\rmX_0,\rvy^{(k)}; \Theta), \quad \forall k\in\sN,
\end{equation}
Because of the continuity of $f$ and the compactness of $[0, 1]^{N\times P}$, we can take limits \wrt $k$ on both sides, which gives
\begin{equation}
    f(\rmA(\rmX_0,\rvy^*); \Theta) = f(\rmX_0,\rvy^*; \Theta) = f_\Theta^*(\rmX_0).
\end{equation}
This shows the self-coherence of the convergent point between $f_\Theta^*$ and $\rvy^*$.

\paragraph{Generalization to novel inter-part states}
Suppose the network has seen $(\rmX,\rvy_{\text{gt}})$ in the training set and learned $f_\Theta(\rmX,\rvy_{\text{gt}};\Theta_*) = \rvy_{\text{gt}}$ with the loss function in Eq. \ref{eq:training} minimized to zero.
Now apply an inter-part transformation $\rmA\in\SEthree^P$ and test the network on $(\rmX', \rvy_{\text{gt}}) := \rmA(\rmX,\rvy_{\text{gt}})$.
We would like to show $f_\Theta^*(\rmX') = f_\Theta^*(\rmX) = \rvy_{\text{gt}}$.
First of all, we know that $\rvy_{\text{gt}}$ itself is a fixed point of $f(\rmX', \cdot;\Theta)$, which is due to the equivariance \wrt $\rvy_{\text{gt}}$ at training time:
\begin{equation}
    f(\rmX',\rvy_{\text{gt}}; \Theta) = f(\rmA(\rmX,\rvy_{\text{gt}}); \Theta) = f(\rmX,\rvy_{\text{gt}}; \Theta).
\end{equation}
Now if $f$ is a contractive function on $\rvy$, we have the \textbf{uniqueness} of Banach fixed-points and thus $\rvy_{\text{gt}}$ is \textbf{the only} fixed point for $f(\rmX', \cdot;\Theta)$, implying that the iterations in Eq. \ref{eq:banach_iter} must converge to $\rvy_{\text{gt}}$.
Equivariance of the per-step iteration and the final convergent state are illustrated in Fig. \ref{fig:method_iters} right.

In less ideal cases when the training loss is not zero but a small error $\varepsilon$, we view it as the distance $\|\rvy^{(1)} - \rvy^{(0)}\|$ between the ground-truth $\rvy^{(0)} = \rvy_{\text{gt}}$ and the one-step iteration output $\rvy^{(1)}$.
The distance between the actual fixed-point $\rvy^*$ and the ground-truth $\rvy_{\text{gt}}$ is then bounded by
\begin{equation}
    \|\rvy^* - \rvy_{\text{gt}}\|
    = \sum_{k=0}^\infty \|\rvy^{(k+1)} - \rvy^{(k)}\|
    \leq \sum_{k=0}^\infty L^k \|\rvy^{(1)} - \rvy^{(0)}\|
    = \frac{1}{1-L}\varepsilon
\end{equation}
where $L$ is the Lipschitz constant of the network.

\subsection{Part Permutations}
\label{sec:method1:part_permut}

A pointcloud segmentation represented by $\rvy \in [0,1]^{N\times P}$ inherently carries an order between the $P$ parts, which is the assumption in semantic segmentation problems.
But in various real-world scenarios, it is common to encounter situations where multiple disjoint parts are associated with the same semantic label, without any clear or coherent part orderings.
In order to tackle this challenge, we also extend our method to encompass the instance segmentation problem without part orderings.
For simplicity, here we assume that all parts can be permuted together by $\Sym_P$.
In practice, permutations only occur among the parts within each semantic label, for which we can easily substitute the $\Sym_P$ below with its subgroup and the conclusion still holds.
We define an equivalence relation on $[0, 1]^{N\times P}$ by
\begin{equation}
    \rvy_1 \sim \rvy_2 \iff \exists \sigma\in\Sym_P, \st \rvy_1\sigma = \rvy_2,
\end{equation}
which gives us a quotient space $[0,1]^{N\times P} / \Sym_P$ and each equivalent class $\hrvy$ in this quotient space represents an instance segmentation without part ordering.
We can define a metric on $\hat{\rvy}$ by
\begin{equation}
\label{eq:dist_quotient}
    d\left( \hrvy_1, \hrvy_2 \right) := \min_{\sigma\in\Sym_P} \|\rvy_1\sigma - \rvy_2\|,
\end{equation}
which makes $\left([0,1]^{N\times P} / \Sym_P, d\right)$ a complete metric space, on which the Banach fixed-point theorem holds and so are the iterations and convergence properties.
Proofs for the well-definedness of $d$ and the completeness of the quotient space are shown in the \SupplementaryMaterial.
\section{Part-Aware Equivariant Network}
\label{sec:method2:network}

\begin{figure*}[t]
    \centering
    \includegraphics[width=0.99\textwidth]{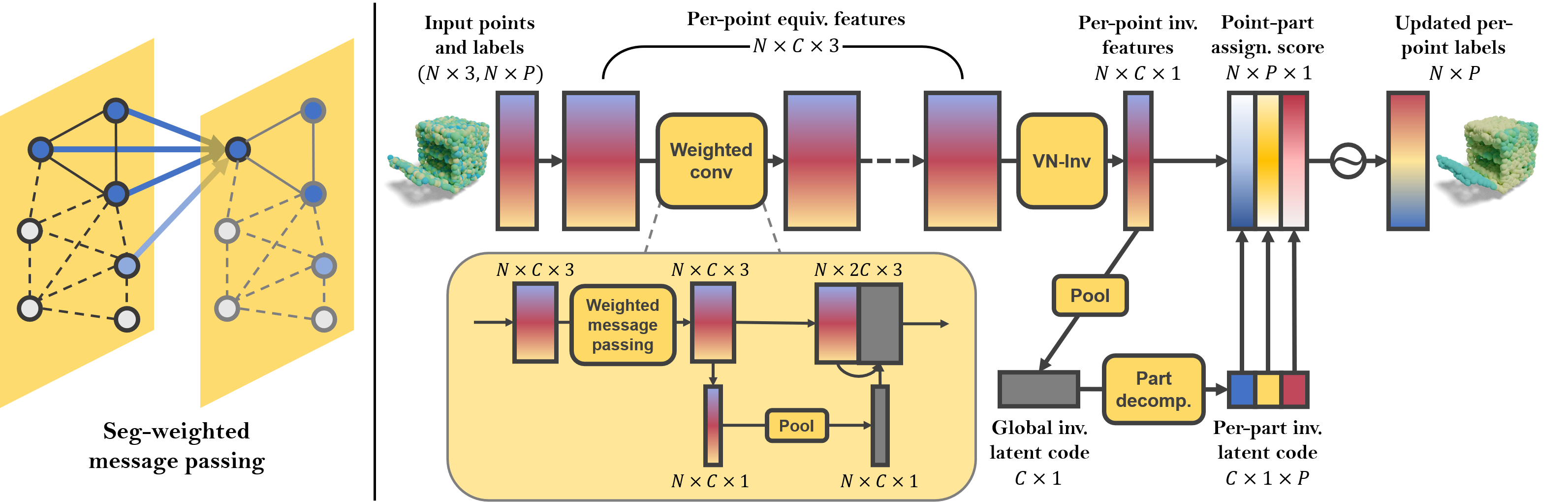}
    \caption{\textbf{Part-aware equivariant network.}
    \textbf{Left:} Segmentation-weighted message passing.
    \textbf{Right:} Overall architecture for segmentation label update.
    Color gradients on the tensors indicate the usage of segmentation labels,
    }
    \label{fig:method_network}
\end{figure*}

In this section, we present our part-aware equivariant network for segmentation label updates (Fig.\ref{fig:method_network}). We utilize a $\SEthree$-equivariant framework called Vector Neurons (VN) \cite{deng2021vector} as the backbone to encode per-point features $\rmV \in \sR^{N\times C\times3}$, ensuring equivariance within each part. Subsequently, these features are transformed into invariant representations, eliminating relative poses to facilitate inter-part information propagation and global aggregations.

\paragraph{Segmentation-weighted message passing}
Key to our part-aware equivariant network is a message-passing module weighted by the input segmentation $\rvy$ (Fig. \ref{fig:method_network} left).
Intuitively, given a point $\rvx_n\in\rmX$ with latent feature $\rmV_n$, for its neighborhood points $\rvx_m\in\gN$, we only allow information propagation between $\rvx_n$ and $\rvx_m$ if they belong to the same part.
When the part segmentation $\rvy$ is a soft mask, we compute the probability $p_{nm}$ of $\rvx_n$ and $\rvx_m$ belonging to the same part by $p_{nm} = \rvy_n\rvy_m^t, \, \rvy_n, \rvy_m \in [0,1]^{1\times P}$ and weight the local message passing with
\begin{equation}
\label{eq:message_passing}
    \rmV_n' = \sum_{m\in\gN} p_{mn}~ \varphi(\rmV_m - \rmV_n, \rmV_n) \bigg/ \sum_{m\in\gN} p_{mn}.
\end{equation}
Proof of inter-part equivariance for this module can be found in the \SupplementaryMaterial.

\paragraph{Network architecture}
Fig. \ref{fig:method_network} right shows the overall network architecture for segmentation label updates.
Given an input pair $(\rmX, \rvy)$, we first extract a per-point $\SEthree$-equivariant local feature $\rmV \in \sR^{N\times C\times3}$ within each part masked by $\rvy$.
$\rmV$ is then passed through a sequence of weighted convolutions with the massage passing defined in Eq. \ref{eq:message_passing}.
To enable efficient inter-part information exchange, after each message-passing layer, we compute a global invariant feature through a per-point VN-invariant layer, followed by a global pooling and a concatenation between per-point and global features.
After the point convolutions, the per-point equivariant features $\rmV_n$ are converted to invariant features $\rmS_n$ with inter-part poses canceled.
We then apply a global pooling on $\rmS_n$ to obtain a global shape code, which is decomposed into $P$ part codes $\rmQ_1, \cdots, \rmQ_P$ as in \cite{chen2019bae}.
Finally, we compute a point-part assignment score for all $(\rmS_n, \rmQ_p)$ pairs (resulting in an $N\times P$ tensor) and apply a softmax activation along its $P$ dimension to obtain the updated segmentation $\rvy'$.
For instance segmentation without part permutations (Sec. \ref{sec:method1:part_permut}), we modify the computation of the part feature $\rmQ$ by replacing the global feature decomposition with point feature grouping based on $\rvy$.

For pointcloud networks with set-invariance across all points, directly restricting the upper bound of the network Lipschitz using weight truncations \cite{virmaux2018lipschitz, gouk2021regularisation, zhang2022rethinking} will greatly harm network expressivity as it limits the output range on each point.
On the other hand, the space of all possible segmentations $[0,1]^{N \times P}$ has extremely high dimensionality, making the computation of Lipschitz regularization losses \cite{terjek2019adversarial} rather inefficient.
Therefore, we do not explicitly constrain the Lipschitz constant of our network. Nevertheless, we confine all operations involving $\rvy$ to local neighborhoods per point, as shown in Eq. \ref{eq:message_passing}. This practical approach generally ensures a small Lipschitz constant for the network in most scenarios.
The stability of fixed-point convergence will be studied in Sec. \ref{sec:exp:ablation}.
\section{Experiments}
\label{sec:exp}

\subsection{Articulated-object part segmentation}
\begin{table*}[t]
\centering
\setlength{\tabcolsep}{2pt}

\scalebox{0.9}{
\begin{tabular}{l|cccc|cccc}
\toprule
Setting
& \multicolumn{4}{c|}{\textbf{Unseen states}}
& \multicolumn{4}{c}{\textbf{Unseen states + unseen instances}} \\
\midrule
Category
& \begin{tabular}[c]{c}Washing\\machine\end{tabular}
& Oven
& \begin{tabular}[c]{c}Eye-\\glasses\end{tabular}
& \begin{tabular}[c]{c}Refrige-\\rator\end{tabular}
& \begin{tabular}[c]{c}Washing\\machine\end{tabular}
& Oven
& \begin{tabular}[c]{c}Eye-\\glasses\end{tabular}
& \begin{tabular}[c]{c}Refrige-\\rator\end{tabular}
\\
\midrule
PointNet \cite{qi2017pointnet}
& 46.18\tiny{~$\pm$~3.84} & 44.08\tiny{~$\pm$~8.97} & 38.96\tiny{~$\pm$18.41} & 38.37\tiny{~$\pm$12,32}
& 46.15\tiny{~$\pm$~3.18} & 45.29\tiny{~$\pm$~9.54} & 39.21\tiny{~$\pm$17.14} & 39.00\tiny{~$\pm$12.13} \\
DGCNN \cite{wang2019dynamic}
& 46.78\tiny{~$\pm$~4.37} & 44.30\tiny{~$\pm$10.84} & 33.35\tiny{~$\pm$20.96} & 39.70\tiny{~$\pm$14.05}
& 46.60\tiny{~$\pm$~4.03} & 46.69\tiny{~$\pm$12.76} & 34.96\tiny{~$\pm$22.14} & 40.13\tiny{~$\pm$13.81} \\
VNN \cite{deng2021vector}
& 47.35\tiny{~$\pm$~1.93} & 53.64\tiny{~$\pm$13.27} & 57.08\tiny{~$\pm$15.52} & 52.36\tiny{~$\pm$11.39}
& 47.09\tiny{~$\pm$~1.84} & 51.01\tiny{~$\pm$11.09} & 62.98\tiny{~$\pm$11.09} & 48.49\tiny{~$\pm$11.58} \\
\textbf{Ours}
& \textbf{82.32}\tiny{~$\pm$15.08} & \textbf{81.91}\tiny{~$\pm$10.81} & \textbf{77.78}\tiny{~$\pm$14.45} & \textbf{77.26}\tiny{~$\pm$~7.79}
& \textbf{84.99}\tiny{~$\pm$11.76} & \textbf{82.84}\tiny{~$\pm$~8.13} & \textbf{78.65}\tiny{~$\pm$10.36} & \textbf{73.93}\tiny{~$\pm$14.08} \\
\bottomrule
\end{tabular}
}
\\
\caption{\textbf{Shape2Motion results.} Numbers are segmentation IoU multiplied by 100.
}
\label{tab:exp:s2m}
\end{table*}

\begin{figure*}[t]
    \centering
    \includegraphics[width=0.99\linewidth]{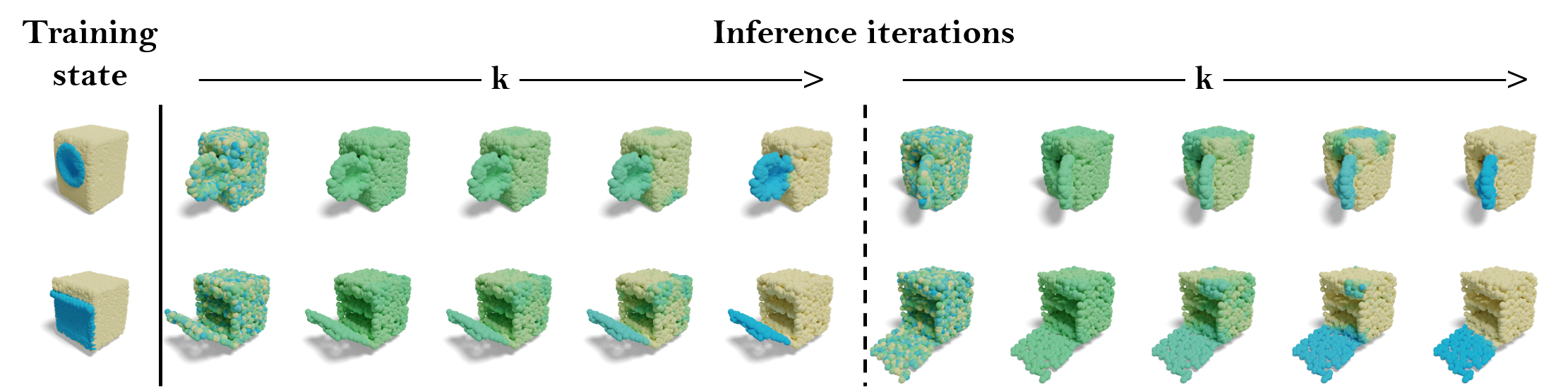}
    \caption{
    \textbf{Banach iterations on Shape2Motion.} The network is trained on rest-state objects (left) and tested on novel articulation states. 
    }
    \label{fig:exp_s2m_iter}
\end{figure*}
\begin{figure*}[t]
    \centering
    \includegraphics[width=0.99\linewidth]{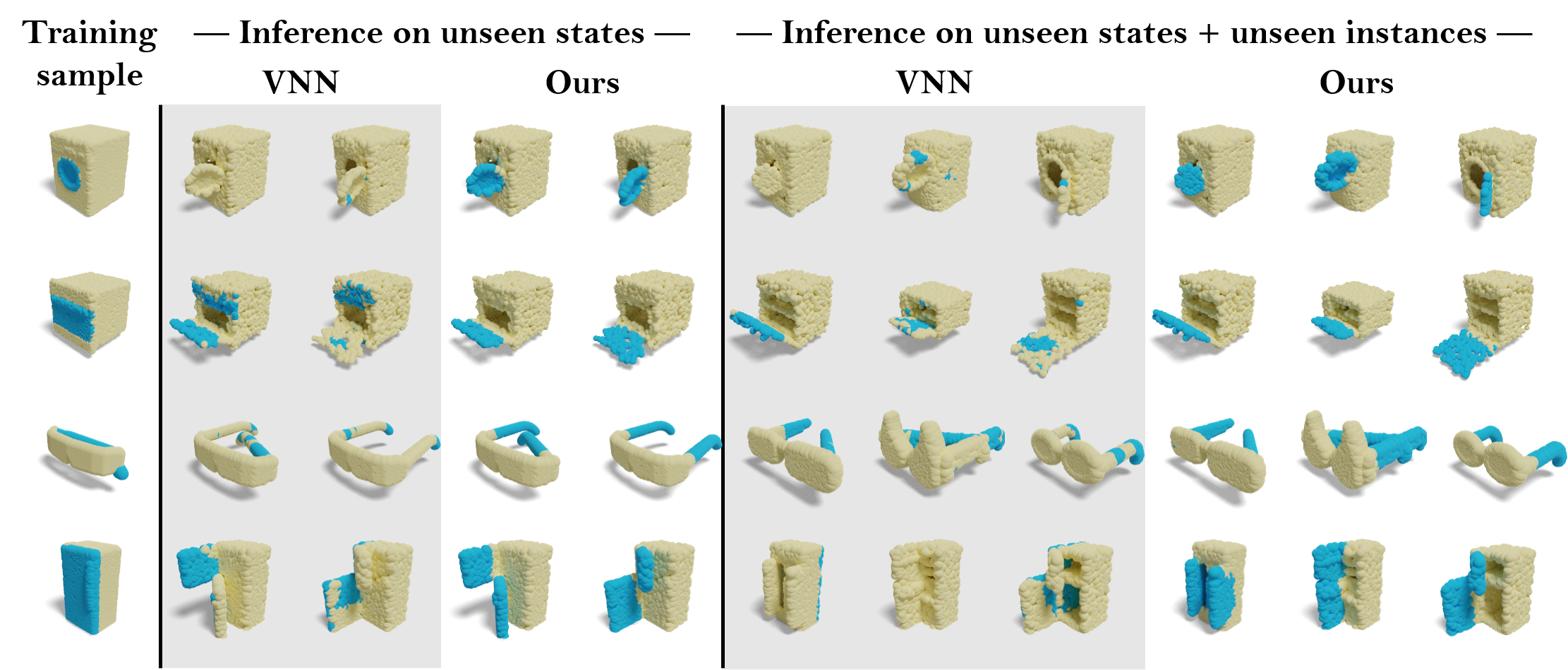}
    \caption{
    \textbf{Shape2Motion part segmentation results.}
    \emph{Global poses are aligned here for visualization but we do not assume it in our inference.}
    }
    \label{fig:exp_s2m_test}
\end{figure*}

We begin by evaluating our method on articulated-object part segmentation using on four categories of the Shape2Motion dataset \cite{wang2019shape2motion}: washing machine, oven, eyeglasses, and refrigerator.
To demonstrate the generalizability of our approach, we train our model on objects in a single rest state, such as ovens with closed doors (as depicted in Fig. \ref{fig:exp_s2m_test}, left). This training setup aligns with many synthetic datasets featuring static shapes \cite{chang2015shapenet, mo2019partnet}. Subsequently, we assess the performance of our model on articulated states, where global and inter-part pose transformations are applied, replicating real-world scenarios.
We evaluate our model on both unseen articulation states of the training instances (Tab. \ref{tab:exp:s2m}, left) and on unseen instances (Tab. \ref{tab:exp:s2m}, right). In both cases, the joint angles are uniformly sampled from the motion range for each category.
We compare our network to the most widely adopted object-level part segmentation networks PointNet \cite{qi2017pointnet} and DGCNN \cite{wang2019dynamic} without equivariance, plus VNN \cite{deng2021vector} with global $\SEthree$-equivariance.
All IoUs are computed for semantic parts. In the case of eyeglasses and refrigerators, which have two parts sharing the same semantic label, we train our network with three motion part inputs but output two semantic part labels. This is feasible because our weighted message passing (Eq. \ref{eq:message_passing}) is agnostic to the number of parts.

Fig. \ref{fig:exp_s2m_iter} demonstrates our iterative inference process on novel states after training on the rest states.
Fig. \ref{fig:exp_s2m_test} shows our segmentation predictions on the four categories. The inter-part pose transformations can cause significant changes in pointcloud geometry and topology, yet our method can robustly generalize to the unseen articulations with inter-part equivariance.

\subsection{Multi-Object Scans}

\begin{figure}[t]
\begin{minipage}[b]{.34\linewidth}
% ================================================= == %
\centering
\setlength{\tabcolsep}{2pt}
\begin{tabular}{l|l}
\toprule
Method & Seg. IoU \\
\midrule
PointNet++ \cite{qi2017pointnet++}          & 39.4\tiny{$\pm$~7.1} \\
MeteorNet \cite{liu2019meteornet}            & 71.8\tiny{$\pm$~9.7} \\
DeepPart \cite{yi2018deep}                  & 66.3\tiny{$\pm$~17.2} \\
NPP \cite{hayden2020nonparametric}          & 71.6\tiny{$\pm$~7.7} \\
Ward-linkage \cite{ward1963hierarchical}    & 88.6\tiny{$\pm$~5.8} \\
PointGroup \cite{jiang2020pointgroup}       & 72.4\tiny{$\pm$~12.5} \\
MultiBodySync \cite{huang2021multibodysync} & 94.0\tiny{$\pm$~3.1} \\
\textbf{Ours}                               & \textbf{95.5}\tiny{$\pm$~4.6} \\
\bottomrule
\end{tabular}
\\
\captionof{table}{\textbf{DynLab segmentation.} Numbers are segmentation IoU multiplied by 100.
\label{tab:exp:dynlab}
}
% =================================================== %
\end{minipage}
\begin{minipage}[b]{.64\linewidth}
% =================================================== %
\centering
\includegraphics[width=.99\textwidth]{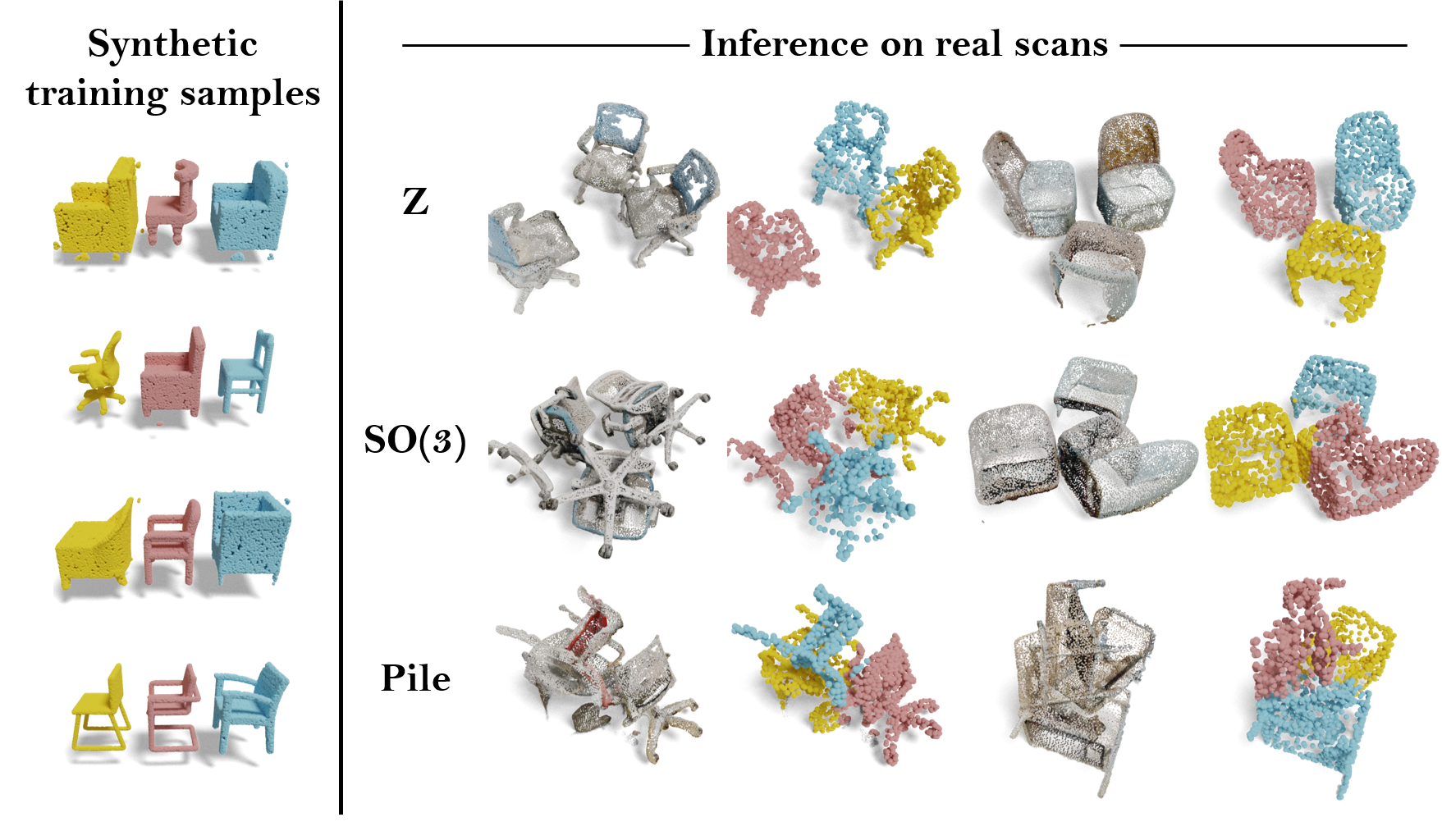}
\caption{\textbf{Chair scan segmentation.}}
\label{fig:exp:chairs}
% =================================================== %
\end{minipage}

\end{figure}

We also test our instance segmentation framework (without part orders) on multi-object scans.

\paragraph{Segmentation transfer on DynLab}
DynLab \cite{huang2021multibodysync} is a collection of scanned laboratory scenes, each with 2-3 rigidly moving solid objects captured under 8 different configurations with object positions randomly changed.
We overfit our model to the first configuration of each scene and apply it to the 7 novel configurations, transfering the segmentation from the first scan to the others via inter-object equivariance.

We compare our method to a variety of multi-body segmentations methods, including motion-based co-segmentation methods (DeepPart \cite{yi2018deep}, NPP \cite{hayden2020nonparametric}, MultiBodySync \cite{huang2021multibodysync}), direct segmentation prediction models (PointNet++ \cite{qi2017pointnet++}, MeteorNet \cite{liu2019meteornet}), geometry-based point grouping (Ward-linkage \cite{ward1963hierarchical}), and pre-trained indoor instance segmentation modules (PointGroup \cite{jiang2020pointgroup}). The baseline setups follow \cite{huang2021multibodysync}.
Tab. \ref{tab:exp:dynlab} shows the segmentation IoUs.
One thing to note is that motion-based co-segmentation can also be viewed as learning inter-object equivariance via Siamese training, yet they cannot reach the performance of equivariance by construction.

\paragraph{Synthetic to real chair scans}
We train our model using a synthetic dataset constructed from clean ShapeNet chair models \cite{chang2015shapenet} with all instances lined up and facing the same direction (Fig. \ref{fig:exp:chairs} left). We then test our model on the real chair scans from \cite{lei2023efem} with diverse scene configurations (Fig. \ref{fig:exp:chairs} right). The configurations range from easy to hard, including: \textbf{Z} (all chairs standing on the floor), \textbf{SO(3)} (chairs laid down), and \textbf{Pile} (chairs piled into clutters). Remarkably, our model with inter-object equivariance demonstrates successful generalization across all these scenarios, even in the most challenging Pile setting with cluttered objects.

\subsection{Ablation Studies}
\label{sec:exp:ablation}

\begin{figure}[t]
\begin{minipage}[b]{.48\linewidth}
% =================================================== %
\centering
\setlength{\tabcolsep}{2pt}
\begin{tabular}{l|cc|c}
\toprule
Method & Equiv. & Pose cano. & Seg. IoU \\
\midrule
PCA           & --         & rot.+trans. & 31.22\tiny{~$\pm$~7.65} \\
VNN           & $\SOthree$ & trans.      & 27.86\tiny{$\pm$~13.07} \\
\textbf{Ours} & $\SEthree$ & --          & \textbf{82.84}\tiny{~$\pm$~8.13} \\
\bottomrule
\end{tabular}
\\
\captionof{table}{\textbf{Ablations of per-part network equivariant operators replaced by pose canonicalization.} Numbers are segmentation IoU multiplied by 100.
\label{tab:exp:ablation}
}
% =================================================== %
\end{minipage}
~
\begin{minipage}[b]{.48\linewidth}
% =================================================== %
\centering
\includegraphics[width=.9\textwidth]{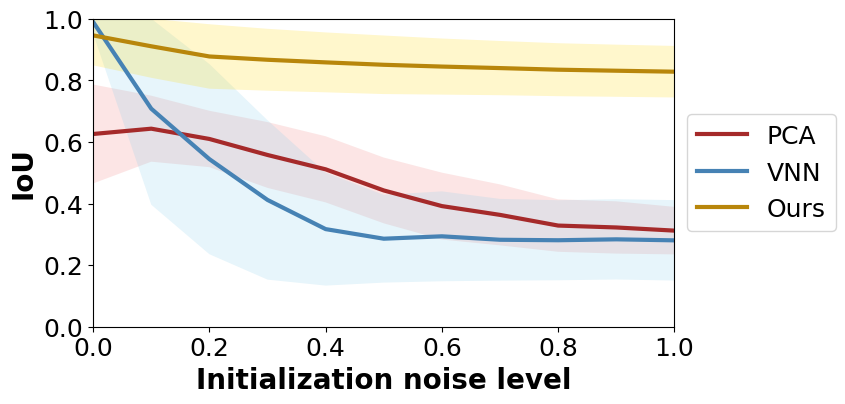}
\caption{\textbf{Model performances with initializations of different noise levels.}}
\label{fig:exp:ablation}
% =================================================== %
\end{minipage}

\end{figure}

We show the importance of our locally confined message passing and $\SEthree$-features (Eq. \ref{eq:message_passing}) to the convergence of test-time iterations.
As opposed to extracting part-level equivariant features, one can also apply pose canonicalizations to each part individually and employ non-equivariant learning methods in the canonical space. For $\SEthree$-transformations with a translation component in $\sR^3$ and a rotation component in $\SOthree$, the former can be canonicalized by subtracting the part centers and the latter by per-part PCA.
Thus we compare our full model with two ablated versions with \emph{part-level} equivariance replaced by canonicalization: \textbf{PCA}, where we canonicalize both the part translations and rotations and simply employ a non-equivariant DGCNN \cite{wang2019dynamic} for the weighted message passing; \textbf{VNN}, where we only canonicalize the part translations but preserve the $\SOthree$-equivariant structures in the network.

Tab. \ref{tab:exp:ablation} shows the segmentation IoUs of the full model and the ablated versions on the Shape2Motion oven category.
espite canonicalizations being agnostic to per-part $\SEthree$ transformations, they rely on absolute poses and positions, which breaks the locality of network operations, resulting in a significant increase in the Lipschitz constant of the network.
Consequently, when any local equivariant operator is replaced by canonicalization, the network performance experiences a drastic drop.

To further examine the convergence ranges, we test the three models under different $\rvy^{(0)}$ initializations by adding random noises $\xi\sim\gU(0,1)$ to $\rvy_{\text{gt}}$ according to $\rvy^{(0)} = (1-\alpha)\rvy_{\text{gt}} + \alpha\xi, \, \alpha \in [0,1]$.
Fig. \ref{fig:exp:ablation} shows the performance changes of the three models with gradually increased noise levels on $\rvy^{(0)}$.
The PCA rotation canonicalization has a theoretically unbounded Lipschitz constant, making the network unable to converge stably even within a small neighborhood of $\rvy_{\text{gt}}$.
The $\SOthree$-equivariant VNN with translation canonicalization perfectly overfits to the ground-truth fixed point but exhibits a rapid decline in performance as the initial $\rvy^{(0)}$ deviates from $\rvy_{\text{gt}}$.
In contrast, our $\SEthree$-equivariant network demonstrates stable performance across different noise levels.

\section{Conclusions}

In this work, we propose \textit{Banana}, which provides both theoretical insights and experimental analysis of inter-part equivariance.
While equivariance is typically examined from an algebraic perspective, we approach it from an analysis standpoint and obtain a strict formulation of inter-part equivariance at the convergent point of an iteration sequence.
Based on our theoretical formulation, we propose a novel framework that co-evolves segmentation labels and per-part $\SEthree$-equivariance, showing strong generalization under both global and per-part transformations even when these transformations result in subsequent changes to the pointcloud geometry or topology.
Experimentally, we show that our model can generalize from rest-state articulated objects to unseen articulations, and from synthetic toy multi-object scenes to real scans with diverse object configurations.

\paragraph{Limitations and future work}
While our local segmentation-weighted message passing reduces the Lipschitz constant of the network, practically providing stability during test-time iterations, it is important to note that it is not explicitly bounded or regularized. Lipschitz bounds and regularizations for set-invariant networks without compromising network expressivity would be an interesting and important problem for future study.
In addition, we study the full $\SEthree^P$ action on multi-body systems, but in many real-world scenarios, different parts or objects may not move independently due to physical constraints.
Limiting the $\SEthree^P$-action to its subset of physically plausible motions can potentially increase our feature-embedding conciseness and learning efficiency.

\paragraph{Broader Impacts} Our study focuses on the general 3D geometry and its equivariance properties. Specifically, we investigate everyday appliances in common scenes, for which we do not anticipate any direct negative social impact. However, we acknowledge that our work could potentially be misused for harmful purposes. On the other hand, our research also has potential applications in areas that can benefit society, such as parsing proteins in medical science and building home robots for senior care.

\begin{ack}
We gratefully acknowledge the following grants: a TRI University 2.0 grant, a Vannevar Bush Faculty Fellowship, and a gift from the Adobe Corporation awarded to Stanford University; and NSF FRR 2220868, NSF IIS-RI 2212433, NSF TRIPODS 1934960, NSF CPS 2038873 awarded to the University of Pennsylvania.
\end{ack}

\renewcommand{\thesection}{S.\arabic{section}}
\renewcommand{\thetable}{S\arabic{table}}
\renewcommand{\thefigure}{S\arabic{figure}}

\section{Proof for Part Permutations (Sec 3.4)}
\label{appen:proof_permute}

In this section, we show that $[0,1]^{N\times P} / \Sym_P$ is a Banach space with metric
\begin{equation*}
    d\left( \hrvy_1, \hrvy_2 \right) := \min_{\sigma\in\Sym_P} \|\rvy_1\sigma - \rvy_2\|.
\end{equation*}

\subsection{Well-Definednes of $d$}
As $d$ is defined on equivalent classes, we need to show the well-definedness of $d$ in that $d$ is independent of the choice of representatives.
Suppose $\rvy_1 = \rvy_1',\, \rvy_2 = \rvy_2'$, then by the definition of equivalent classes, $\exists \sigma_1,\sigma_2\in\Sym_P,\, \st \rvy_1\sigma_1 = \rvy_1',\, \rvy_2\sigma_2 = \rvy_2'$.
On one side,
\begin{align*}
    d(\hrvy_1,\hrvy_2)
    &= \min_{\sigma\in\Sym_P} \|\rvy_1\sigma - \rvy_2\|
    \leq \|\rvy_1\sigma_1\sigma_2^{-1} - \rvy_2\| \\
    &= \|\rvy_1\sigma_1 - \rvy_2\sigma_2\|
    = d(\hrvy_1',\hrvy_2').
\end{align*}
Similarly, we also have $d(\hrvy_1',\hrvy_2') \leq d(\hrvy_1,\hrvy_2)$, and thus $d(\hrvy_1,\hrvy_2) = d(\hrvy_1',\hrvy_2')$.

\subsection{$d$ as a Metric}
To show that $d$ is a metric, we need to show its \textbf{positivity}, \textbf{symmetry}, and \textbf{triangle inequality}.

\paragraph{Positivity}
Suppose $\hrvy_1 \neq \hrvy_2$. Then $\forall \sigma\in\Sym_P, \rvy_1\sigma \neq \rvy_2$. This implies
\begin{equation*}
    d(\hrvy_1,\hrvy_2) = \min_{\sigma\in\Sym_P} \| \rvy_1\sigma - \rvy_2 \| > 0
\end{equation*}
as $\Sym_P$ is a finite set and taking minimum over it strictly preserves inequality.
Similarly, we can show that $d(\hrvy,\hrvy) \geq 0$.
Together with
\begin{equation*}
    d(\hrvy, \hrvy) = \min_{\sigma\in\Sym_P} \| \rvy\sigma - \rvy \|
    \leq \| \rvy\mathbf{1} - \rvy \| = \| \rvy - \rvy \| = 0,
\end{equation*}
we get $d(\hrvy,\hrvy) = 0$.

\paragraph{Symmetry}
For any $\hrvy_1,\hrvy_2$, we have
\begin{align*}
    d(\hrvy_1, \hrvy_2)
    &= \min_{\sigma\in\Sym_P} \| \rvy_1\sigma - \rvy_2 \| \\
    &= \min_{\sigma\in\Sym_P} \| \rvy_1\sigma\sigma^{-1} - \rvy_2\sigma^{-1} \| \\
    &= \min_{\sigma\in\Sym_P} \| \rvy_1 - \rvy_2\sigma^{-1} \| \\
    &= \min_{\sigma\in\Sym_P} \| \rvy_2\sigma^{-1} - \rvy_1 \| \\
    &= \min_{\sigma'\in\Sym_P} \| \rvy_2\sigma' - \rvy_1 \|
    = d(\hrvy_2, \hrvy_1).
\end{align*}
Here we use the symmetry of the standard Euclidean norm $\|\cdot\|$, the $l_2$-distance perseverance of permutation matrices, and the fact that $\sigma\in\Sym_P \iff \sigma^{-1}\in\Sym_P$ due to the group structure of $\Sym_P$.

\paragraph{Triangle inequality}
\begin{align*}
    d(\hrvy_1,\hrvy_3)
    &= \min_{\sigma\in\Sym_P} \| \rvy_1\sigma - \rvy_3 \|
    = \min_{\sigma\in\Sym_P} \| \rvy_1\sigma - \rvy_2\sigma' + \rvy_2\sigma' - \rvy_3 \|,   \quad \forall \sigma'\in\Sym_P \\
    &= \min_{\sigma'\in\Sym_P} \min_{\sigma\in\Sym_P} \| \rvy_1\sigma - \rvy_2\sigma' + \rvy_2\sigma' - \rvy_3 \| \\
    &\leq \min_{\sigma',\sigma\in\Sym_P} \| \rvy_1\sigma - \rvy_2\sigma' \| + \| \rvy_2\sigma' - \rvy_3 \| \\
    &= \min_{\sigma',\sigma\in\Sym_P} \| \rvy_1\sigma\sigma'^{-1} - \rvy_2\sigma'\sigma'^{-1} \| + \| \rvy_2\sigma' - \rvy_3 \| \\
    &= \min_{\sigma',\sigma''\in\Sym_P} \| \rvy_1\sigma'' - \rvy_2 \| + \| \rvy_2\sigma' - \rvy_3 \| \\
    &= \min_{\sigma''\in\Sym_P} \| \rvy_1\sigma'' - \rvy_2 \| + \min_{\sigma''\in\Sym_P} \| \rvy_2\sigma' - \rvy_3 \| \\
    &= d(\hrvy_1,\hrvy_2) + d(\hrvy_2,\hrvy_3).
\end{align*}

\subsection{Completeness of $\left( [0,1]^{N\times P} / \Sym_P, d \right)$}
To show the completeness of this space, we need to show that every Cauchy sequence converges to a point in this space.
Suppose $\{\hrvy_i: i\in\sN\}$ is a Cauchy sequence satisfying $\forall \varepsilon>0,\, \exists N>0,\, \st\, \forall i,j> M,\, d(\hrvy_i,\hrvy_j)<\varepsilon$.
Let $\{\rvy_i: i\in\sN\}$ be a sequence of arbitrarily selected representatives from each $\hrvy_i$, it is a bounded sequence in $[0,1]^{N\times P}$, and thus have a convergent subsequence $\{\rvy_{i_k}: k\in\sN\}$ with limit $\lim_{k\to\infty} \rvy_{i_k} = \rvy^* \in [0,1]^{N\times P}$ by the Bolzano–Weierstrass theorem.

We would like to show that $\hrvy^* \in [0,1]^{N\times P} / \Sym_P$ is the limit point of $\{\hrvy_i\}$ using proof by contradiction.
Suppose it is not, then $\exists \varepsilon_0>0, \st \forall M>0, \exists j_0 > M, \st d(\hrvy_{j_0},\hrvy^*) \geq \varepsilon_0$.
By the convergence of $\{\rvy_{i_k}\}$, $\exists M_1$, \st $\forall i_k \geq k > M_1$, $\|\rvy_{i_k} - \rvy^*\| < \varepsilon_0 / 3$.
On the other hand, because $\{\hrvy_i\}$ is Cauchy, $\exists M_2$, \st, $\forall i,j>M_2$, $d(\hrvy_i,\hrvy_j) < \varepsilon_0 / 3$.
Now let $M = \max(M_1, M_2)$, then for $i_k, j_0 > M$, we have
\begin{equation*}
    \varepsilon_0
    \leq d(\hrvy_{j_0},\hrvy^*)
    \leq d(\hrvy_{j_0}, \hrvy_{i_k}) + d(\hrvy_{i_k},\hrvy^*)
    < \varepsilon_0/3 + \|\rvy_{i_k} - \rvy^*\|
    < \varepsilon / 3 + \varepsilon / 3.
\end{equation*}
A contradiction!
\section{Proof for Network Equivariance (Sec. 4)}
\label{appen:proof_equi}

Suppose $\rvy$ is binary, when $\rvx_n,\rvx_m$ belong to different parts ($p(m) \neq p(n)$), $\sum_{p\in P}\rvy_{np}\rvy_{mp} = 0$.
The weighted message passing can be written as
\begin{align*}
    \rmV_n'
    &= \sum_{m\in\gN} \rvy_n\rvy_m^t ~\varphi(\rmV_m - \rmV_n, \rmV_n) \bigg / \sum_{m\in\gN} \rvy_n\rvy_m^t \\
    &= \sum_{m\in\gN} \sum_{p\in P}\rvy_{np}\rvy_{mp} ~\varphi(\rmV_m - \rmV_n, \rmV_n) \bigg / \sum_{m\in\gN} \rvy_n\rvy_m^t \\
    &= \sum_{p(m)=p(n)} ~\varphi(\rmV_m - \rmV_n, \rmV_n) \bigg / \sum_{m\in\gN} \rvy_n\rvy_m^t.
\end{align*}
Now apply a transformation $\rmA = (\rmT_1,\cdots,\rmT_P) \in \SEthree^P$ to the input per-point features $\rmV$, it transforms $\rmV_n, \rmV_m$ into
\begin{equation*}
    \rmV_n \mapsto \sum_{p=1}^P \rvy_{np}(\rmV_n\rmR_p + \rvt_p), \quad
    \rmV_m \mapsto \sum_{p=1}^P \rvy_{mp}(\rmV_m\rmR_p + \rvt_p).
\end{equation*}
When $p(m) = p(n)$, $\sum_{p=1}^P (\rvy_{mp} - \rvy_{np}) = 0$, and the right-hand side of the above equation then turns into
\begin{align*}
    &\sum_{p(m)=p(n)} ~\varphi\bigg(
    \sum_{p=1}^P \rvy_{mp}(\rmV_m\rmR_p + \rvt_p) -
    \sum_{p=1}^P \rvy_{np}(\rmV_n\rmR_p + \rvt_p),
    \sum_{p=1}^P \rvy_{np}(\rmV_n\rmR_p + \rvt_p)
    \bigg) \bigg / \sum_{m\in\gN} \rvy_n\rvy_m^t \\
    &= \sum_{p(m)=p(n)} ~\varphi\bigg(
    \sum_{p=1}^P \rvy_{np}((\rmV_m - \rmV_n)\rmR_p + \rvt_p) +
    \sum_{p=1}^P (\rvy_{mp} - \rvy_{np})(\rmV_m\rmR_p + \rvt_p), \\
    &\qquad\qquad\qquad\qquad\qquad\qquad
    \sum_{p=1}^P \rvy_{np}(\rmV_n\rmR_p + \rvt_p)
    \bigg) \bigg / \sum_{m\in\gN} \rvy_n\rvy_m^t \\
    &= \sum_{p(m)=p(n)} ~\varphi\bigg(
    \sum_{p=1}^P \rvy_{np}((\rmV_m - \rmV_n)\rmR_p + \rvt_p), 
    \sum_{p=1}^P \rvy_{np}(\rmV_n\rmR_p + \rvt_p)
    \bigg) \bigg / \sum_{m\in\gN} \rvy_n\rvy_m^t \\
    &= \sum_{p(m)=p(n)} ~\varphi\bigg(
    ((\rmV_m - \rmV_n)\rmR_{p(n)} + \rvt_{p(n)}), 
    (\rmV_n\rmR_{p(n)} + \rvt_{p(n)})
    \bigg) \bigg / \sum_{m\in\gN} \rvy_n\rvy_m^t.
\end{align*}
By the part-wise equivariance of $\varphi$, this is equal to $\rmV_n'\rmR_{p(n)} + \rvt_{p(n)}$.

\section{Network and Training Details}

In our network, we use 2 weighted message-passing layers to extract the per-point $\SEthree$-equivariant features, followed by 4 weighted message-passing layers with global invariant feature concatenation. Output features are passed through a 3-layer MLP to obtain the final segmentation labels. All network layers have latent dimension 128.
For the neighborhood search in our message-passing layers, we use a ball query with radius $r=0.3$ with maximum $k=40$ points.
We use an Adam optimizer with an initial learning rate of 0.001.
All networks are trained on one single NVIDIA Titan RTX 24GB GPU.

{
\small
\bibliographystyle{plain}
\bibliography{egbib}

\begin{thebibliography}{10}

\bibitem{agarwal2018banach}
Praveen Agarwal, Mohamed Jleli, Bessem Samet, Praveen Agarwal, Mohamed Jleli,
  and Bessem Samet.
\newblock Banach contraction principle and applications.
\newblock {\em Fixed Point Theory in Metric Spaces: Recent Advances and
  Applications}, pages 1--23, 2018.

\bibitem{aronsson2022homogeneous}
Jimmy Aronsson.
\newblock Homogeneous vector bundles and g-equivariant convolutional neural
  networks.
\newblock {\em Sampling Theory, Signal Processing, and Data Analysis},
  20(2):10, 2022.

\bibitem{assaad2022vn}
Serge Assaad, Carlton Downey, Rami Al-Rfou, Nigamaa Nayakanti, and Ben Sapp.
\newblock Vn-transformer: Rotation-equivariant attention for vector neurons.
\newblock {\em arXiv preprint arXiv:2206.04176}, 2022.

\bibitem{baddeley1992working}
Alan Baddeley.
\newblock Working memory.
\newblock {\em Science}, 255(5044):556--559, 1992.

\bibitem{baddeley2012working}
Alan Baddeley.
\newblock Working memory: Theories, models, and controversies.
\newblock {\em Annual review of psychology}, 63:1--29, 2012.

\bibitem{bansal2022end}
Arpit Bansal, Avi Schwarzschild, Eitan Borgnia, Zeyad Emam, Furong Huang, Micah
  Goldblum, and Tom Goldstein.
\newblock End-to-end algorithm synthesis with recurrent networks: Logical
  extrapolation without overthinking.
\newblock {\em arXiv preprint arXiv:2202.05826}, 2022.

\bibitem{baur2021slim}
Stefan~Andreas Baur, David~Josef Emmerichs, Frank Moosmann, Peter Pinggera,
  Bj{\"o}rn Ommer, and Andreas Geiger.
\newblock Slim: Self-supervised lidar scene flow and motion segmentation.
\newblock In {\em Proceedings of the IEEE/CVF International Conference on
  Computer Vision}, pages 13126--13136, 2021.

\bibitem{behl2018pointflownet}
Aseem Behl, Despoina Paschalidou, Simon Donn{\'e}, and Andreas Geiger.
\newblock Pointflownet: Learning representations for 3d scene flow estimation
  from point clouds.
\newblock {\em arXiv preprint arXiv:1806.02170}, 2, 2018.

\bibitem{chang2015shapenet}
Angel~X Chang, Thomas Funkhouser, Leonidas Guibas, Pat Hanrahan, Qixing Huang,
  Zimo Li, Silvio Savarese, Manolis Savva, Shuran Song, Hao Su, et~al.
\newblock Shapenet: An information-rich 3d model repository.
\newblock {\em arXiv preprint arXiv:1512.03012}, 2015.

\bibitem{chatzipantazis2022se}
Evangelos Chatzipantazis, Stefanos Pertigkiozoglou, Edgar Dobriban, and Kostas
  Daniilidis.
\newblock Se (3)-equivariant attention networks for shape reconstruction in
  function space.
\newblock {\em arXiv preprint arXiv:2204.02394}, 2022.

\bibitem{chen2021equivariant}
Haiwei Chen, Shichen Liu, Weikai Chen, Hao Li, and Randall Hill.
\newblock Equivariant point network for 3d point cloud analysis.
\newblock In {\em Proceedings of the IEEE/CVF conference on computer vision and
  pattern recognition}, pages 14514--14523, 2021.

\bibitem{chen20224dcontrast}
Yujin Chen, Matthias Nie{\ss}ner, and Angela Dai.
\newblock 4dcontrast: Contrastive learning with dynamic correspondences for 3d
  scene understanding.
\newblock In {\em Computer Vision--ECCV 2022: 17th European Conference, Tel
  Aviv, Israel, October 23--27, 2022, Proceedings, Part XXXII}, pages 543--560.
  Springer, 2022.

\bibitem{chen2019bae}
Zhiqin Chen, Kangxue Yin, Matthew Fisher, Siddhartha Chaudhuri, and Hao Zhang.
\newblock Bae-net: Branched autoencoder for shape co-segmentation.
\newblock In {\em Proceedings of the IEEE/CVF International Conference on
  Computer Vision}, pages 8490--8499, 2019.

\bibitem{cohen2019general}
Taco~S Cohen, Mario Geiger, and Maurice Weiler.
\newblock A general theory of equivariant cnns on homogeneous spaces.
\newblock {\em Advances in neural information processing systems}, 32, 2019.

\bibitem{deng2021vector}
Congyue Deng, Or~Litany, Yueqi Duan, Adrien Poulenard, Andrea Tagliasacchi, and
  Leonidas~J Guibas.
\newblock Vector neurons: A general framework for so (3)-equivariant networks.
\newblock In {\em Proceedings of the IEEE/CVF International Conference on
  Computer Vision}, pages 12200--12209, 2021.

\bibitem{dinh2014nice}
Laurent Dinh, David Krueger, and Yoshua Bengio.
\newblock Nice: Non-linear independent components estimation.
\newblock {\em arXiv preprint arXiv:1410.8516}, 2014.

\bibitem{dinh2016density}
Laurent Dinh, Jascha Sohl-Dickstein, and Samy Bengio.
\newblock Density estimation using real nvp.
\newblock {\em arXiv preprint arXiv:1605.08803}, 2016.

\bibitem{fu2022robust}
Jiahui Fu, Yilun Du, Kurran Singh, Joshua~B Tenenbaum, and John~J Leonard.
\newblock Robust change detection based on neural descriptor fields.
\newblock In {\em 2022 IEEE/RSJ International Conference on Intelligent Robots
  and Systems (IROS)}, pages 2817--2824. IEEE, 2022.

\bibitem{fuchs2020se}
Fabian Fuchs, Daniel Worrall, Volker Fischer, and Max Welling.
\newblock Se (3)-transformers: 3d roto-translation equivariant attention
  networks.
\newblock {\em Advances in Neural Information Processing Systems},
  33:1970--1981, 2020.

\bibitem{girshick2015fast}
Ross Girshick.
\newblock Fast r-cnn.
\newblock In {\em Proceedings of the IEEE international conference on computer
  vision}, pages 1440--1448, 2015.

\bibitem{girshick2014rich}
Ross Girshick, Jeff Donahue, Trevor Darrell, and Jitendra Malik.
\newblock Rich feature hierarchies for accurate object detection and semantic
  segmentation.
\newblock In {\em Proceedings of the IEEE conference on computer vision and
  pattern recognition}, pages 580--587, 2014.

\bibitem{goller1996learning}
Christoph Goller and Andreas Kuchler.
\newblock Learning task-dependent distributed representations by
  backpropagation through structure.
\newblock In {\em Proceedings of International Conference on Neural Networks
  (ICNN'96)}, volume~1, pages 347--352. IEEE, 1996.

\bibitem{gouk2021regularisation}
Henry Gouk, Eibe Frank, Bernhard Pfahringer, and Michael~J Cree.
\newblock Regularisation of neural networks by enforcing lipschitz continuity.
\newblock {\em Machine Learning}, 110:393--416, 2021.

\bibitem{hayden2020nonparametric}
David~S Hayden, Jason Pacheco, and John~W Fisher.
\newblock Nonparametric object and parts modeling with lie group dynamics.
\newblock In {\em Proceedings of the IEEE/CVF Conference on Computer Vision and
  Pattern Recognition}, pages 7426--7435, 2020.

\bibitem{he2017mask}
Kaiming He, Georgia Gkioxari, Piotr Doll{\'a}r, and Ross Girshick.
\newblock Mask r-cnn.
\newblock In {\em Proceedings of the IEEE international conference on computer
  vision}, pages 2961--2969, 2017.

\bibitem{higuera2022neural}
Carolina Higuera, Siyuan Dong, Byron Boots, and Mustafa Mukadam.
\newblock Neural contact fields: Tracking extrinsic contact with tactile
  sensing.
\newblock {\em arXiv preprint arXiv:2210.09297}, 2022.

\bibitem{ho2020denoising}
Jonathan Ho, Ajay Jain, and Pieter Abbeel.
\newblock Denoising diffusion probabilistic models.
\newblock {\em Advances in Neural Information Processing Systems},
  33:6840--6851, 2020.

\bibitem{hochreiter1997long}
Sepp Hochreiter and J{\"u}rgen Schmidhuber.
\newblock Long short-term memory.
\newblock {\em Neural computation}, 9(8):1735--1780, 1997.

\bibitem{huang2021multibodysync}
Jiahui Huang, He~Wang, Tolga Birdal, Minhyuk Sung, Federica Arrigoni, Shi-Min
  Hu, and Leonidas~J Guibas.
\newblock Multibodysync: Multi-body segmentation and motion estimation via 3d
  scan synchronization.
\newblock In {\em Proceedings of the IEEE/CVF Conference on Computer Vision and
  Pattern Recognition}, pages 7108--7118, 2021.

\bibitem{huang2022dynamic}
Shengyu Huang, Zan Gojcic, Jiahui Huang, Andreas Wieser, and Konrad Schindler.
\newblock Dynamic 3d scene analysis by point cloud accumulation.
\newblock In {\em Computer Vision--ECCV 2022: 17th European Conference, Tel
  Aviv, Israel, October 23--27, 2022, Proceedings, Part XXXVIII}, pages
  674--690. Springer, 2022.

\bibitem{jain2021screwnet}
Ajinkya Jain, Rudolf Lioutikov, Caleb Chuck, and Scott Niekum.
\newblock Screwnet: Category-independent articulation model estimation from
  depth images using screw theory.
\newblock In {\em 2021 IEEE International Conference on Robotics and Automation
  (ICRA)}, pages 13670--13677. IEEE, 2021.

\bibitem{jiang2020pointgroup}
Li~Jiang, Hengshuang Zhao, Shaoshuai Shi, Shu Liu, Chi-Wing Fu, and Jiaya Jia.
\newblock Pointgroup: Dual-set point grouping for 3d instance segmentation.
\newblock In {\em Proceedings of the IEEE/CVF conference on computer vision and
  Pattern recognition}, pages 4867--4876, 2020.

\bibitem{jiang2022ditto}
Zhenyu Jiang, Cheng-Chun Hsu, and Yuke Zhu.
\newblock Ditto: Building digital twins of articulated objects from
  interaction.
\newblock In {\em Proceedings of the IEEE/CVF Conference on Computer Vision and
  Pattern Recognition}, pages 5616--5626, 2022.

\bibitem{kar2019evidence}
Kohitij Kar, Jonas Kubilius, Kailyn Schmidt, Elias~B Issa, and James~J DiCarlo.
\newblock Evidence that recurrent circuits are critical to the ventral
  stream’s execution of core object recognition behavior.
\newblock {\em Nature neuroscience}, 22(6):974--983, 2019.

\bibitem{katzir2022shape}
Oren Katzir, Dani Lischinski, and Daniel Cohen-Or.
\newblock Shape-pose disentanglement using se (3)-equivariant vector neurons.
\newblock In {\em Computer Vision--ECCV 2022: 17th European Conference, Tel
  Aviv, Israel, October 23--27, 2022, Proceedings, Part III}, pages 468--484.
  Springer, 2022.

\bibitem{kawana2021unsupervised}
Yuki Kawana, Yusuke Mukuta, and Tatsuya Harada.
\newblock Unsupervised pose-aware part decomposition for 3d articulated
  objects.
\newblock {\em arXiv preprint arXiv:2110.04411}, 2021.

\bibitem{kingma2018glow}
Durk~P Kingma and Prafulla Dhariwal.
\newblock Glow: Generative flow with invertible 1x1 convolutions.
\newblock {\em Advances in neural information processing systems}, 31, 2018.

\bibitem{kondor2018generalization}
Risi Kondor and Shubhendu Trivedi.
\newblock On the generalization of equivariance and convolution in neural
  networks to the action of compact groups.
\newblock In {\em International Conference on Machine Learning}, pages
  2747--2755. PMLR, 2018.

\bibitem{Lei2022CaDeX}
Jiahui Lei and Kostas Daniilidis.
\newblock Cadex: Learning canonical deformation coordinate space for dynamic
  surface representation via neural homeomorphism.
\newblock In {\em Proceedings of the IEEE/CVF Conference on Computer Vision and
  Pattern Recognition}, 2022.

\bibitem{lei2023efem}
Jiahui Lei, Congyue Deng, Karl Schmeckpeper, Leonidas Guibas, and Kostas
  Daniilidis.
\newblock Efem: Equivariant neural field expectation maximization for 3d object
  segmentation without scene supervision.
\newblock {\em arXiv preprint arXiv:2303.15440}, 2023.

\bibitem{li2022directed}
Jiahan Li, Shitong Luo, Congyue Deng, Chaoran Cheng, Jiaqi Guan, Leonidas
  Guibas, Jian Peng, and Jianzhu Ma.
\newblock Directed weight neural networks for protein structure representation
  learning.
\newblock {\em arXiv preprint arXiv:2201.13299}, 2022.

\bibitem{li2021self}
Ruibo Li, Guosheng Lin, and Lihua Xie.
\newblock Self-point-flow: Self-supervised scene flow estimation from point
  clouds with optimal transport and random walk.
\newblock In {\em Proceedings of the IEEE/CVF conference on computer vision and
  pattern recognition}, pages 15577--15586, 2021.

\bibitem{li2021leveraging}
Xiaolong Li, Yijia Weng, Li~Yi, Leonidas~J Guibas, A~Abbott, Shuran Song, and
  He~Wang.
\newblock Leveraging se (3) equivariance for self-supervised category-level
  object pose estimation from point clouds.
\newblock {\em Advances in Neural Information Processing Systems},
  34:15370--15381, 2021.

\bibitem{liao2016bridging}
Qianli Liao and Tomaso Poggio.
\newblock Bridging the gaps between residual learning, recurrent neural
  networks and visual cortex.
\newblock {\em arXiv preprint arXiv:1604.03640}, 2016.

\bibitem{lin2022coarse}
Cheng-Wei Lin, Tung-I Chen, Hsin-Ying Lee, Wen-Chin Chen, and Winston~H Hsu.
\newblock Coarse-to-fine point cloud registration with se (3)-equivariant
  representations.
\newblock {\em arXiv preprint arXiv:2210.02045}, 2022.

\bibitem{liu20173dcnn}
Fangyu Liu, Shuaipeng Li, Liqiang Zhang, Chenghu Zhou, Rongtian Ye, Yuebin
  Wang, and Jiwen Lu.
\newblock 3dcnn-dqn-rnn: A deep reinforcement learning framework for semantic
  parsing of large-scale 3d point clouds.
\newblock In {\em Proceedings of the IEEE international conference on computer
  vision}, pages 5678--5687, 2017.

\bibitem{liu2019flownet3d}
Xingyu Liu, Charles~R Qi, and Leonidas~J Guibas.
\newblock Flownet3d: Learning scene flow in 3d point clouds.
\newblock In {\em Proceedings of the IEEE/CVF Conference on Computer Vision and
  Pattern Recognition}, pages 529--537, 2019.

\bibitem{liu2019meteornet}
Xingyu Liu, Mengyuan Yan, and Jeannette Bohg.
\newblock Meteornet: Deep learning on dynamic 3d point cloud sequences.
\newblock In {\em Proceedings of the IEEE/CVF International Conference on
  Computer Vision}, pages 9246--9255, 2019.

\bibitem{liu2023self}
Xueyi Liu, Ji~Zhang, Ruizhen Hu, Haibin Huang, He~Wang, and Li~Yi.
\newblock Self-supervised category-level articulated object pose estimation
  with part-level se (3) equivariance.
\newblock {\em arXiv preprint arXiv:2302.14268}, 2023.

\bibitem{luo2021diffusion}
Shitong Luo and Wei Hu.
\newblock Diffusion probabilistic models for 3d point cloud generation.
\newblock In {\em Proceedings of the IEEE/CVF Conference on Computer Vision and
  Pattern Recognition}, pages 2837--2845, 2021.

\bibitem{mo2019partnet}
Kaichun Mo, Shilin Zhu, Angel~X Chang, Li~Yi, Subarna Tripathi, Leonidas~J
  Guibas, and Hao Su.
\newblock Partnet: A large-scale benchmark for fine-grained and hierarchical
  part-level 3d object understanding.
\newblock In {\em Proceedings of the IEEE/CVF conference on computer vision and
  pattern recognition}, pages 909--918, 2019.

\bibitem{mu2021sdf}
Jiteng Mu, Weichao Qiu, Adam Kortylewski, Alan Yuille, Nuno Vasconcelos, and
  Xiaolong Wang.
\newblock A-sdf: Learning disentangled signed distance functions for
  articulated shape representation.
\newblock In {\em Proceedings of the IEEE/CVF International Conference on
  Computer Vision}, pages 13001--13011, 2021.

\bibitem{pan2022so}
Haoran Pan, Jun Zhou, Yuanpeng Liu, Xuequan Lu, Weiming Wang, Xuefeng Yan, and
  Mingqiang Wei.
\newblock So (3)-pose: So (3)-equivariance learning for 6d object pose
  estimation.
\newblock {\em arXiv preprint arXiv:2208.08338}, 2022.

\bibitem{poulenard2021functional}
Adrien Poulenard and Leonidas~J Guibas.
\newblock A functional approach to rotation equivariant non-linearities for
  tensor field networks.
\newblock In {\em Proceedings of the IEEE/CVF Conference on Computer Vision and
  Pattern Recognition}, pages 13174--13183, 2021.

\bibitem{puy2020flot}
Gilles Puy, Alexandre Boulch, and Renaud Marlet.
\newblock Flot: Scene flow on point clouds guided by optimal transport.
\newblock In {\em Computer Vision--ECCV 2020: 16th European Conference,
  Glasgow, UK, August 23--28, 2020, Proceedings, Part XXVIII}, pages 527--544.
  Springer, 2020.

\bibitem{qi2017pointnet}
Charles~R Qi, Hao Su, Kaichun Mo, and Leonidas~J Guibas.
\newblock Pointnet: Deep learning on point sets for 3d classification and
  segmentation.
\newblock In {\em Proceedings of the IEEE conference on computer vision and
  pattern recognition}, pages 652--660, 2017.

\bibitem{qi2017pointnet++}
Charles~Ruizhongtai Qi, Li~Yi, Hao Su, and Leonidas~J Guibas.
\newblock Pointnet++: Deep hierarchical feature learning on point sets in a
  metric space.
\newblock {\em Advances in neural information processing systems}, 30, 2017.

\bibitem{rezende2015variational}
Danilo Rezende and Shakir Mohamed.
\newblock Variational inference with normalizing flows.
\newblock In {\em International conference on machine learning}, pages
  1530--1538. PMLR, 2015.

\bibitem{ryu2022equivariant}
Hyunwoo Ryu, Jeong-Hoon Lee, Hong-in Lee, and Jongeun Choi.
\newblock Equivariant descriptor fields: Se (3)-equivariant energy-based models
  for end-to-end visual robotic manipulation learning.
\newblock {\em arXiv preprint arXiv:2206.08321}, 2022.

\bibitem{sajnani2022condor}
Rahul Sajnani, Adrien Poulenard, Jivitesh Jain, Radhika Dua, Leonidas~J Guibas,
  and Srinath Sridhar.
\newblock Condor: Self-supervised canonicalization of 3d pose for partial
  shapes.
\newblock In {\em Proceedings of the IEEE/CVF Conference on Computer Vision and
  Pattern Recognition}, pages 16969--16979, 2022.

\bibitem{schwarzschild2021can}
Avi Schwarzschild, Eitan Borgnia, Arjun Gupta, Furong Huang, Uzi Vishkin, Micah
  Goldblum, and Tom Goldstein.
\newblock Can you learn an algorithm? generalizing from easy to hard problems
  with recurrent networks.
\newblock {\em Advances in Neural Information Processing Systems},
  34:6695--6706, 2021.

\bibitem{simeonov2022neural}
Anthony Simeonov, Yilun Du, Andrea Tagliasacchi, Joshua~B Tenenbaum, Alberto
  Rodriguez, Pulkit Agrawal, and Vincent Sitzmann.
\newblock Neural descriptor fields: Se (3)-equivariant object representations
  for manipulation.
\newblock In {\em 2022 International Conference on Robotics and Automation
  (ICRA)}, pages 6394--6400. IEEE, 2022.

\bibitem{song2020denoising}
Jiaming Song, Chenlin Meng, and Stefano Ermon.
\newblock Denoising diffusion implicit models.
\newblock {\em arXiv preprint arXiv:2010.02502}, 2020.

\bibitem{song2019generative}
Yang Song and Stefano Ermon.
\newblock Generative modeling by estimating gradients of the data distribution.
\newblock {\em Advances in neural information processing systems}, 32, 2019.

\bibitem{song2020score}
Yang Song, Jascha Sohl-Dickstein, Diederik~P Kingma, Abhishek Kumar, Stefano
  Ermon, and Ben Poole.
\newblock Score-based generative modeling through stochastic differential
  equations.
\newblock {\em arXiv preprint arXiv:2011.13456}, 2020.

\bibitem{song2022ogc}
Ziyang Song and Bo~Yang.
\newblock Ogc: Unsupervised 3d object segmentation from rigid dynamics of point
  clouds.
\newblock {\em arXiv preprint arXiv:2210.04458}, 2022.

\bibitem{tang2023diffuscene}
Jiapeng Tang, Yinyu Nie, Lev Markhasin, Angela Dai, Justus Thies, and Matthias
  Nie{\ss}ner.
\newblock Diffuscene: Scene graph denoising diffusion probabilistic model for
  generative indoor scene synthesis.
\newblock {\em arXiv preprint arXiv:2303.14207}, 2023.

\bibitem{terjek2019adversarial}
D{\'a}vid Terj{\'e}k.
\newblock Adversarial lipschitz regularization.
\newblock {\em arXiv preprint arXiv:1907.05681}, 2019.

\bibitem{thomas2021self}
Hugues Thomas, Ben Agro, Mona Gridseth, Jian Zhang, and Timothy~D Barfoot.
\newblock Self-supervised learning of lidar segmentation for autonomous indoor
  navigation.
\newblock In {\em 2021 IEEE International Conference on Robotics and Automation
  (ICRA)}, pages 14047--14053. IEEE, 2021.

\bibitem{thomas2018tensor}
Nathaniel Thomas, Tess Smidt, Steven Kearnes, Lusann Yang, Li~Li, Kai Kohlhoff,
  and Patrick Riley.
\newblock Tensor field networks: Rotation-and translation-equivariant neural
  networks for 3d point clouds.
\newblock {\em arXiv preprint arXiv:1802.08219}, 2018.

\bibitem{virmaux2018lipschitz}
Aladin Virmaux and Kevin Scaman.
\newblock Lipschitz regularity of deep neural networks: analysis and efficient
  estimation.
\newblock {\em Advances in Neural Information Processing Systems}, 31, 2018.

\bibitem{wang2016cnn}
Jiang Wang, Yi~Yang, Junhua Mao, Zhiheng Huang, Chang Huang, and Wei Xu.
\newblock Cnn-rnn: A unified framework for multi-label image classification.
\newblock In {\em Proceedings of the IEEE conference on computer vision and
  pattern recognition}, pages 2285--2294, 2016.

\bibitem{wang2019shape2motion}
Xiaogang Wang, Bin Zhou, Yahao Shi, Xiaowu Chen, Qinping Zhao, and Kai Xu.
\newblock Shape2motion: Joint analysis of motion parts and attributes from 3d
  shapes.
\newblock In {\em Proceedings of the IEEE/CVF Conference on Computer Vision and
  Pattern Recognition}, pages 8876--8884, 2019.

\bibitem{wang2019dynamic}
Yue Wang, Yongbin Sun, Ziwei Liu, Sanjay~E Sarma, Michael~M Bronstein, and
  Justin~M Solomon.
\newblock Dynamic graph cnn for learning on point clouds.
\newblock {\em Acm Transactions On Graphics (tog)}, 38(5):1--12, 2019.

\bibitem{wang20224d}
Yuqi Wang, Yuntao Chen, and ZHAO-XIANG ZHANG.
\newblock 4d unsupervised object discovery.
\newblock {\em Advances in Neural Information Processing Systems},
  35:35563--35575, 2022.

\bibitem{ward1963hierarchical}
Joe~H Ward~Jr.
\newblock Hierarchical grouping to optimize an objective function.
\newblock {\em Journal of the American statistical association},
  58(301):236--244, 1963.

\bibitem{wei2023lego}
Qiuhong~Anna Wei, Sijie Ding, Jeong~Joon Park, Rahul Sajnani, Adrien Poulenard,
  Srinath Sridhar, and Leonidas Guibas.
\newblock Lego-net: Learning regular rearrangements of objects in rooms.
\newblock {\em arXiv e-prints}, pages arXiv--2301, 2023.

\bibitem{weiler2021coordinate}
Maurice Weiler, Patrick Forr{\'e}, Erik Verlinde, and Max Welling.
\newblock Coordinate independent convolutional networks--isometry and gauge
  equivariant convolutions on riemannian manifolds.
\newblock {\em arXiv preprint arXiv:2106.06020}, 2021.

\bibitem{weng2022neural}
Thomas Weng, David Held, Franziska Meier, and Mustafa Mukadam.
\newblock Neural grasp distance fields for robot manipulation.
\newblock {\em arXiv preprint arXiv:2211.02647}, 2022.

\bibitem{xu2022unified}
Yinshuang Xu, Jiahui Lei, Edgar Dobriban, and Kostas Daniilidis.
\newblock Unified fourier-based kernel and nonlinearity design for equivariant
  networks on homogeneous spaces.
\newblock In {\em International Conference on Machine Learning}, pages
  24596--24614. PMLR, 2022.

\bibitem{xue2022useek}
Zhengrong Xue, Zhecheng Yuan, Jiashun Wang, Xueqian Wang, Yang Gao, and Huazhe
  Xu.
\newblock Useek: Unsupervised se (3)-equivariant 3d keypoints for generalizable
  manipulation.
\newblock {\em arXiv preprint arXiv:2209.13864}, 2022.

\bibitem{yang2019pointflow}
Guandao Yang, Xun Huang, Zekun Hao, Ming-Yu Liu, Serge Belongie, and Bharath
  Hariharan.
\newblock Pointflow: 3d point cloud generation with continuous normalizing
  flows.
\newblock In {\em Proceedings of the IEEE/CVF international conference on
  computer vision}, pages 4541--4550, 2019.

\bibitem{ye20183d}
Xiaoqing Ye, Jiamao Li, Hexiao Huang, Liang Du, and Xiaolin Zhang.
\newblock 3d recurrent neural networks with context fusion for point cloud
  semantic segmentation.
\newblock In {\em Proceedings of the European conference on computer vision
  (ECCV)}, pages 403--417, 2018.

\bibitem{yi2018deep}
Li~Yi, Haibin Huang, Difan Liu, Evangelos Kalogerakis, Hao Su, and Leonidas
  Guibas.
\newblock Deep part induction from articulated object pairs.
\newblock {\em arXiv preprint arXiv:1809.07417}, 2018.

\bibitem{yu2022rotationally}
Hong-Xing Yu, Jiajun Wu, and Li~Yi.
\newblock Rotationally equivariant 3d object detection.
\newblock In {\em Proceedings of the IEEE/CVF Conference on Computer Vision and
  Pattern Recognition}, pages 1456--1464, 2022.

\bibitem{zhang2022rethinking}
Bohang Zhang, Du~Jiang, Di~He, and Liwei Wang.
\newblock Rethinking lipschitz neural networks and certified robustness: A
  boolean function perspective.
\newblock In {\em Advances in Neural Information Processing Systems}, 2022.

\bibitem{zhou20213d}
Linqi Zhou, Yilun Du, and Jiajun Wu.
\newblock 3d shape generation and completion through point-voxel diffusion.
\newblock In {\em Proceedings of the IEEE/CVF International Conference on
  Computer Vision}, pages 5826--5835, 2021.

\bibitem{zhu2022correspondence}
Minghan Zhu, Maani Ghaffari, and Huei Peng.
\newblock Correspondence-free point cloud registration with so (3)-equivariant
  implicit shape representations.
\newblock In {\em Conference on Robot Learning}, pages 1412--1422. PMLR, 2022.

\end{thebibliography}
}

%%%%%%%%%%%%%%%%%%%%%%%%%%%%%%%%%%%%%%%%%%%%%%%%%%%%%%%%%%%%

\end{document}